%% file: main.tex
\pdfoutput=1

\documentclass[11pt]{article}

\usepackage{acl}

\usepackage{times}
\usepackage{latexsym}
\usepackage{booktabs}
\usepackage{booktabs}
\usepackage{amsfonts}
\usepackage{nicefrac}
\usepackage{microtype}
\usepackage{todonotes}
\usepackage{listings}
\usepackage{algorithm}
\usepackage{algpseudocode}
\usepackage{soul}
\usepackage{tabularx}
\usepackage{pifont}
\usepackage{xcolor}
\usepackage{graphicx}
\usepackage{subcaption}
\usepackage{hyperref}
\usepackage{comment}
\usepackage{tcolorbox}
\usepackage{nicematrix}

\usepackage{colortbl}
\usepackage{xcolor}
\usepackage{makecell}

\usepackage[normalem]{ulem}

\input{macro}

\usepackage[T1]{fontenc}

\usepackage[utf8]{inputenc}

\usepackage{microtype}

\usepackage{algpseudocode,algorithm}
\usepackage{hyperref}
\usepackage{multirow}
\usepackage{graphicx}
\usepackage{pdfpages}
\usepackage{url}
\usepackage{xcolor}
\usepackage{epsfig}
\usepackage{adjustbox}
\usepackage{amsfonts}
\usepackage{amsmath}
\usepackage{amssymb}
\usepackage{booktabs} 
\usepackage{comment}
\newcommand\coco{\raisebox{-1pt}{\includegraphics[width=1em]{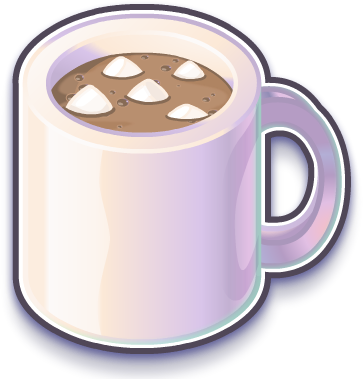}}}

\title{
\coco~\textsc{CoCoA}: CBT-based Conversational Counseling Agent using Memory Specialized in Cognitive Distortions and Dynamic Prompt}
\author{
  Suyeon Lee$^{1}$\thanks{~~Equal contribution} \qquad \textbf{Jieun Kang}$^{1*}$ \qquad  \textbf{Harim Kim}$^{2}$\\
   \textbf{Kyoung-Mee Chung}$^{2}$\qquad \textbf{Dongha Lee}$^{1}$ \qquad \textbf{Jinyoung Yeo}$^{1}$ \\ 
Department of Artificial Intelligence, Yonsei University$^{1}$ \\
Department of Psychology, Yonsei University$^{2}$ \\
  \texttt{\{isuy.groot, likemika605\}@gmail.com}\\
  \texttt{\{harim.k, kmchung, donalee, jinyeo\}@yonsei.ac.kr}
}

\begin{document}
\maketitle
\input{0_abs}
\input{1_intro}

\input{3_method}
\input{4_exp_new}
\input{2_rel}
\input{6_conc}

\bibliography{anthology,custom}
\bibliographystyle{acl_natbib}

\newpage
\appendix
\input{8_appendix}

\end{document}

%% file: macro.tex
\definecolor{mypink1}{rgb}{0.858, 0.188, 0.478}

\definecolor{olive}{rgb}{0.39, 0.875, 0.12}

\definecolor{yellow-green}{rgb}{0.3, 0.5, 0.0}

\definecolor{lightblue}{RGB}{224,236,247}
\definecolor{deepblue}{RGB}{9,46,107}
\definecolor{defaultcolor}{gray}{0.9}

%% file: 0_abs.tex
\begin{abstract}
    The demand for conversational agents that provide mental health care is consistently increasing. In this work, we develop a psychological counseling agent, referred to as \textbf{CoCoA} that applies Cognitive Behavioral Therapy (CBT) techniques to identify and address cognitive distortions inherent in the client's statements. Specifically, we construct a memory system to efficiently manage information necessary for counseling while extracting high-level insights about the client from their utterances. Additionally, to ensure that the counseling agent generates appropriate responses, we introduce a dynamic prompting to flexibly apply CBT techniques and facilitate the appropriate retrieval of information. We conducted dialogues between \textbf{CoCoA} and characters from Character.ai, creating a dataset for evaluation. Then, we asked GPT to evaluate the constructed counseling dataset and our model demonstrated a statistically significant difference from other models.
\end{abstract}

%% file: 1_intro.tex
\section{Introduction}

In today's world, an increasing number of individuals are expressing worries about their mental health~\citep{KFFCNN2022}. Consequently, there is a higher demand for psychological counseling to address these concerns. However, some argue that they are unable to access appropriate mental healthcare due to factors such as cost and societal stigma~\citep{COVID2022}. In such cases, a conversational agent for psychological counseling can be a valuable alternative for those who are hesitant to seek traditional mental healthcare. Furthermore, it can effectively address the rising demand for mental healthcare by enhancing accessibility and convenience.


Recent studies have developed methods to improve mental state by leveraging conversational agents such as empathetic dialogue generation~\citep{rashkin2018towards} and Emotional Support Conversation~\citep{wang2023enhancing}. However, the above studies focus on providing emotional support, without advancing towards resolving fundamental psychological issues~\citep{liu2021towards}. To address this, we propose a counselor agent based on LLMs that provides psychological counseling by utilizing Cognitive Behavioral Therapy (CBT) techniques. CBT techniques aim to identify and transform cognitive distortions, the root causes of negative emotions. 

To provide human-like counseling using CBT techniques, we suggest several methodologies. 
Counselors need to extract relevant information for response generation from the dialogues with clients. To efficiently extract, store, and retrieve high-level information from the client's utterances, we have structured our memory into two components: \textit{Basic Memory} for storing personal information learned through counseling, and \textit{Cognitive Distortion Memory} for storing identified cognitive distortions from the utterances.
Also, Counselors plan the cognitive distortion to address and select the CBT technique and corresponding stage for resolution based on the client's utterances before responding. We assemble prompts dynamically, enabling our counseling agent to generate responses after undergoing this process. Consequently, we guide our agent to go through the intermediate steps before generating counseling utterances to perform chain-of-thought.

To evaluate our agent's performance, we conduct counseling sessions between our agent \textbf{CoCoA}, and generative agents that imitate well-known personalities. We selected simulacra of 8 renowned figures such as Van Gogh or Jay Gatsby through the Character.ai\footnote{\href{https://beta.character.ai/}{https://beta.character.ai/}} service. Details regarding these personas can be found in \autoref{sec:appendixc}. Then, we instruct GPT-3.5 to evaluate the constructed dataset based on five criteria.

%% file: 3_method.tex
\begin{figure*}
\begin{center}
\includegraphics[width=1\linewidth]{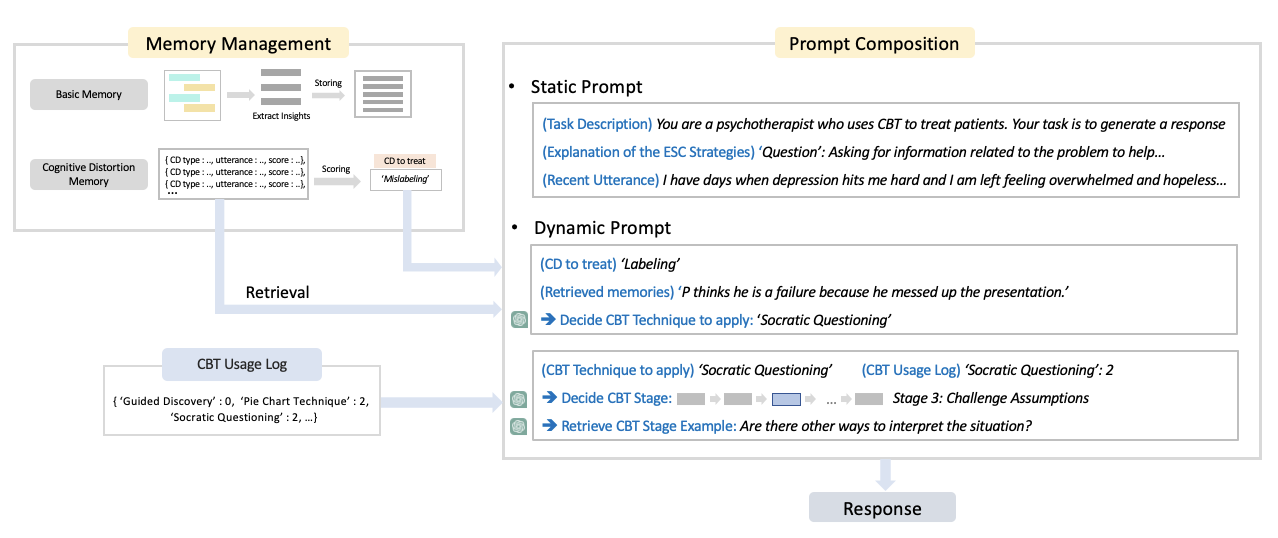}

\end{center}
\caption{CoCoA agent architecture}
\label{fig:short}
\end{figure*}

\section{LLM as a Counselor : \textbf{CoCoA}}
\subsection{Overview}
In this section, we present the Counselor Agent, referred to as \textbf{CoCoA}. As shown in Figure 1, our framework consists of two components: 1) memory management, and 2) prompt composition. When constructing prompts, the required information is retrieved from the memory and used accordingly. The constructed prompt is served as input to the LLM, denoted as $f_{LLM}$. The detailed algorithm, including the architecture through which \textbf{CoCoA} operates, can be found in \autoref{sec:appendixa}.

\noindent\textbf{Memory Management}. The agent stores the information gathered during counseling sessions in memory to provide appropriate responses. Each time client utterances are received, the system draws insights from them and stores insights in the \textit{Basic Memory}. If any cognitive distortions are identified during the conversation, they are saved in the \textit{Cognitive Distortion Memory}.

\noindent\textbf{Prompt Composition}. 
We structure prompts into static $S$ and dynamic $D$ prompts. The static prompt remains unchanged as the conversation progresses, providing a consistent reference. On the other hand, the dynamic prompt is customized for each turn, retrieving knowledge from external sources to generate contextually relevant prompts that fit the current situation. This approach is adopted to flexibly apply Cognitive Behavioral Therapy (CBT) techniques and facilitate the appropriate retrieval of information required during the technique application.

\subsection{Memory Management}


During counseling sessions, LLM finds information hidden in client utterances and stores them in two types of memory: \textit{Basic Memory} and \textit{Cognitive Distortion Memory}, referred to as \textit{CD Memory}. The agent manages and stores cognitive distortion memory separately from the basic memory to identify core beliefs across multiple episodes and integrate them to resolve cognitive distortions.

\textit{Basic Memory} stores the high-level insights extracted from the utterance, while \textit{CD Memory} stores the information related to cognitive distortion. When utterance is given, LLM analyzes each utterance made by the client to identify any cognitive distortions.
If a cognitive distortion is detected, (1) the cognitive distortion type, (2) the utterance, and (3) a score on a Likert scale from 1 to 5, indicating the severity of the distortion are appended to the \textit{CD Memory}.

After constructing \textit{CD Memory}, the agent determines the cognitive distortion type that will be prioritized for treatment among the existing cognitive distortions. To determine which type of cognitive distortion to treat, it considers three main components: \textit{recency}, \textit{frequency}, and \textit{severity}, as inspired by the \citet{park2023generative}. \textit{Recency} assigns a higher score to the most recently mentioned cognitive error, using an exponential decay function over the number of utterances ago. \textit{Frequency} counts the occurrences of each cognitive distortion. \textit{Severity} is determined by the score from 1 to 5 on a Likert scale for each cognitive distortion stored in the \textit{CD Memory}.
The final score is calculated by normalizing the \textit{recency}, \textit{frequency}, and \textit{severity} scores to a value between 0 and 1, and then combining them using weighted coefficients. Thus the scoring function $\textit{S}$ can be denoted as $
\textit{S}(cd) = \alpha_{\textit{recency}} \times \textit{{recency}} + \alpha_{\textit{frequency}} \times \textit{{frequency}} + \alpha_{\textit{severity}} \times \textit{{severity}}$. In our implementation, we have set all $\alpha$s to 1. This process can be denoted as follows:
\begin{equation}
cd^* = \underset{cd  \in CD}{\text{argmax}}  \textit{ S}(cd)
\end{equation}


\subsection{Prompt Composition}
\subsubsection{Static Prompt}
The static prompt consists of a task description $Task$, an explanation of the strategies of the ESC technique $ESC$, and the recent utterances $U$. Thus, the static prompt can be denoted as $ Static = Task + ESC + U$.

\subsubsection{Dynamic Prompt}

The Dynamic prompt is composed of three main processes: (1) determining the optimal CBT technique, (2) specifying the CBT stage to apply, and (3) generating responses based on these determinations.

\noindent\textbf{CBT Technique Determination} 
The system retrieves memories related to $cd^*$ from both \textit{Basic Memory} and \textit{CD Memory}. In this process, we use the Contriever\citep{lei2023unsupervised} that retrieves memories by embedding memories and calculating the similarity between them. The CBT technique to apply, $t^*$,  is selected from the given list of CBT techniques $T$ based on the retrieved memories, $M_{r}$. This process can be denoted as follows:
\begin{eqnarray}
M_r \sim Contriever(\cdot|M_{B}, M_{CD}) \\
 \Rightarrow t^* \sim f_{LLM}(\cdot|M_r, T)
\end{eqnarray}

where $\Rightarrow$ indicates a sequential generation of variable output.

\noindent\textbf{CBT Stage Determination} In the subsequent step, the agent is tasked with selecting a specific stage $Stage$ from the various stages of the chosen CBT technique $t^*$. First, the stages to be taken in the chosen CBT technique are retrieved. Subsequently, the stage of CBT technique to apply is determined by referencing the $Log$ indicating the current progress of CBT technique stages applied in the conversation. Furthermore, the example of an appropriate statement for the current stage is generated simultaneously. This process can be represented as: 
\begin{equation}
Stage, Example \sim f_{LLM}(\cdot|t^*,Log)
\end{equation}

\noindent\textbf{Response Generation} The dynamic prompt is structured with the selected CBT technique, the applicable stage of CBT, and appropriate utterance examples. The dynamic prompt is then integrated with the static prompt to form the input for the LLM, resulting in the generation of a response. This procedure can be illustrated as:
\begin{eqnarray}
Dynamic \sim t^* + Stage + Example \\
Response \sim f_{LLM}(\cdot|Static, Dynamic)
\end{eqnarray}

%% file: 4_exp_new.tex
\section{Experiment}

\begin{figure*}
\begin{center}
\includegraphics[width=0.9\linewidth]{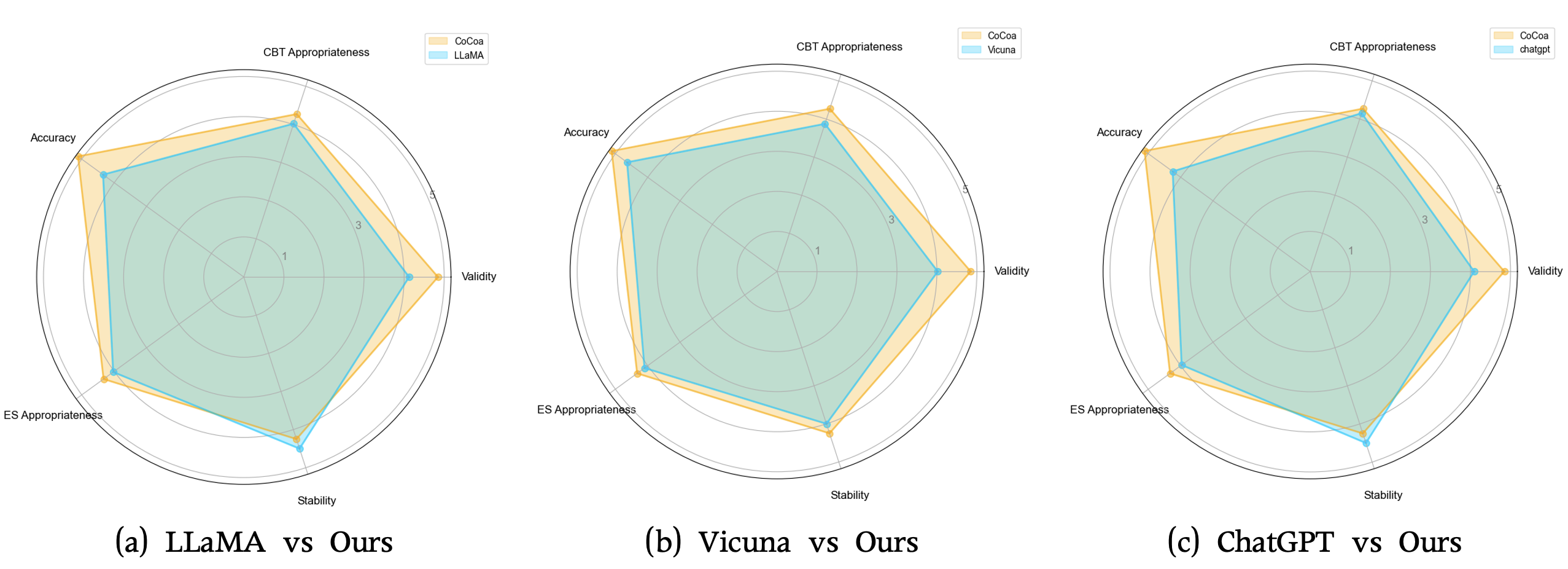}
\end{center}
\caption{Evaluation results}
\label{fig:evals}
\end{figure*}

\subsection{Response Generation Evaluation}
In order to evaluate how well \textbf{CoCoA} performs as a counselor, it is necessary to engage in the conversations between \textbf{CoCoA} and the interlocutor and compare its conversation dataset with those of other LLMs.
To meet this end, we facilitated conversations between the counselor agents and the interlocutor agent, who acts as a client, to generate counseling conversation datasets. Subsequently, we asked \texttt{GPT-3.5-turbo} to rate the constructed dataset with each of the five criteria using a scale of 0 to 6.


\subsection{Evaluation Dataset Construction} 
We opted to use a character.ai\footnote{https://beta.character.ai/} chatbot service to play the role of the client. Character.ai is a conversational agent service that allows you to have a conversation with simulacra created about famous people. We sampled 8 simulacra from character.ai and engaged in conversation with them. The description of the chosen character is provided in Appendix \ref{sec:appendixb}. We compare our models with three baselines: \texttt{GPT-3.5-turbo}, \texttt{LLAMA-2-7B-chat} and \texttt{Vicuna-7B}.

\subsection{Evaluation Criteria}
The evaluation criteria consist of five elements derived from \citet{blackburn2001revised} and \citet{shao2023character}. Three of these elements are related to the validity, appropriateness, and accuracy of the CBT technique application, one is related to the execution of emotionally supportive dialogue, and the other one is about the stability of the model. Each criterion is rated on a scale of 0 to 6.
\begin{itemize}
   \item \textbf{CBT Validity}: The model's ability to choose the optimal technique based on valid reasoning, considering the targeted dysfunctional thoughts and the client's core issues.
   \item \textbf{CBT Appropriateness}: The model should maintain a cooperative attitude, focusing on enabling the client to explore their issues rather than engaging in argumentation or persuasion when using CBT techniques.
   \item \textbf{CBT Accuracy}: The model should use the selected CBT techniques accurately and proficiently.
   \item \textbf{ES Appropriateness}: The model should engage in conversation that remains within the context of the preceding dialogue or general common sense, offering appropriate empathy and comfort.
   \item \textbf{Stability}: The model should remain stable and consistent during relatively long conversations, even in the face of variations in incremental inputs.
    
\end{itemize}






\subsection{Evaluation Results}
Figure \ref{fig:evals} represents the results of evaluating the outputs of each model based on five criteria and calculating the averages. Compared to \texttt{Vicuna-7B}, our model outperforms in all criteria. It generally exhibits high performance on CBT Validity and Accuracy compared to other models. However, there are not a significant difference in ES Appropriateness or Stability. According to the T-test results, \textbf{CoCoa} exhibited statistically significant differences in criteria other than Stability at the 0.05 level.

During the experiment, while other models were provided with the entire conversation history, \textbf{CoCoa} received only the immediate preceding utterance and retrieved necessary information from its memory. Despite this difference, our model's strong performance can be attributed to the construction of a memory specifically tailored for counseling. In essence, the experimental results demonstrate that our model constructs memory effectively and implements appropriate CBT techniques.

%% file: 2_rel.tex
\section{Related Work}

There have been ongoing efforts to introduce conversational agents (CAs) in mental health. Natural Language Processing (NLP) technology has led to the development of task-oriented healthcare dialogue systems. While the research in the field of Computer Science has focused on the technical aspects, other disciplines have explored the usability of Mental Health CAs in psychiatry and user acceptability\cite{cho2023integrative, safi2020technical}. 
Previous studies have traditionally used rule-based or retrieval-based models\cite{daley2020preliminary}.  However, recent research has utilized Large Language Models to improve dialog generation tasks in Mental Health CAs\cite{das2022conversational}.

%% file: 6_conc.tex
\section{Conclusion}
In this study, we explored the potential of developing a counseling agent by proposing a method to construct a cognitive distortion memory and dynamically structure prompts to generate responses based on CBT techniques. We conducted experiments evaluating the model's responses, which demonstrated excellent performance in CBT accuracy and validity. Due to limited availability of counseling datasets from human clients, we opted to use a character.ai. To address the limitation of character.ai for counseling, future research should develop agents with diverse personas and cognitive distortions to simulate human clients.

%% file: 8_appendix.tex
\section{Algorithm of CoCoA}
\label{sec:appendixa}
\label{sec:appendix}
\makeatletter
\algrenewcommand\ALG@beginalgorithmic{\footnotesize}
\algrenewcommand\algorithmiccomment[2][\normalsize]{{#1\hfill\(\triangleright\) #2}}
\makeatother

\begin{algorithm}
\caption{Memory Management}
\begin{algorithmic}[1]
\item \textbf{Input:} 
\item[] $p_{n}$, Client's \textit{n-th} utterance;
\item[] $f_{LLM}$, LLM Model;
\item \textbf{Output:} 
\item[] $M_{B}$, Basic Memory;
\item[] $M_{CD}$, CD Memory;

\item $M_{B} \leftarrow$ Extract Insight from $p_{n}$
\item \textbf{if} $cd$ in $p_{n}$ \textbf{then}
\item[] $M_{CD}$ $\leftarrow$ ( $cd$, $p_{n}$, $severity$ $score$ )
\item \textbf{endif}
\item \textbf{return} $M_{B}, M_{CD}$
\end{algorithmic}
\end{algorithm}

\makeatletter
\algrenewcommand\ALG@beginalgorithmic{\footnotesize}
\algrenewcommand\algorithmiccomment[2][\normalsize]{{#1\hfill\(\triangleright\) #2}}
\makeatother

\begin{algorithm}
\caption{Prompt Composition}
\begin{algorithmic}[1]
    \item[\textbf{Input:}]
        \item $c_{n}$, Counselor's \textit{n-th} Utterance;
        \item $p_{n}$, Client's \textit{n-th} Utterance;
        \item $Task$, Task Description;
        \item $ESC$, ESC strategies;
        \item $M_{B} = \{b_{1}, b_{2}, \dots, b_{n}\}$, Basic Memory;
        \item $M_{CD} = \{d_{1}, d_{2}, \dots, d_{m}\}$, CD Memory;
        \item $CD = \{cd_{1}, cd_{2}, \dots, cd_{l}\}$, Categories of Cognitive Distortion;
        \item $T = \{t_{1}, t_{2}, \dots, t_{p}\}$, Categories of CBT Technique;
        \item $f_{LLM}$, LLM Model;
        \item $Log$, CBT Usage Log;
    \item[\textbf{Output:}]
        \item $c_{n+1}$, Counselor's Next Response;

    \item $U = \{(c_{n}, p_{n})\}$
    \item $\textit{Static} \leftarrow Task  +  ESC  +  U$

    \textbf{if}{$M_{CD} = \phi$}
        \item \textit{Prompt} $\leftarrow$ \textit{Static}
    \textbf{else}
        \item $cd^* \leftarrow \underset{cd \in CD}{\text{argmax}} \textit{ S}(cd)$
        \item $M_{r} \leftarrow \{b_{k} \vert b_{k} \in M_{B}$ and retrieved with $cd^*, U\}$ \\
        \hspace*{4em} $\cup \{d_{k} \vert d_{k} \in M_{CD}$ and retrieved with $cd^*, U\}$
        \item $t^* \leftarrow f_{LLM}(\cdot|M_{r}, T)$
        \item $stage, example \leftarrow f_{LLM}(\cdot|t^*, Log)$
        \item \textit{Prompt} $\leftarrow$ \textit{Static} + \textit{Dynamic}$(t^*, stage, example)$
    \textbf{endif}

    \item $c_{n+1} \leftarrow f_{LLM}(\textit{Prompt}, U)$
    \item \textbf{return} $c_{n+1}$
\end{algorithmic}
\end{algorithm}

\section{Cognitive Behavior Therapy (CBT) Technique}
\label{sec:appendixb}
\subsection{Overview}
CBT is a psychological therapeutic approach that explores cognitive factors in mental disorders and suggests effective intervention methods~\cite{beck1979cognitive}. Individuals perceive the world through automatic thoughts stemming from core beliefs. Negative core beliefs often lead to exaggerating or distorting reality, resulting in various cognitive distortions. In the CBT technique, specific treatment strategies are employed to modify and reconstruct cognitive distortions. A detailed definition of each cognitive distortion and the CBT technique are provided below.
\subsection{The Types of Cognitive Distortions}
The various types of cognitive distortions are as
follows. We referred to the types of cognitive distortions and their explanations presented in \citet{sharma2023cognitive}.

\noindent\textbf{All or nothing thinking} Divide experiences into
categories of either black or white (right or wrong)

\noindent\textbf{Overgeneralization} Make quick judgements
about the entire based on a limited part

\noindent\textbf{Mental filtering} Filter out the positive elements
and dwell excessively on the negative aspects of a
situation

\noindent\textbf{Personalization} Attribute external events to
oneself, even when there is no basis for making
such a connection

\noindent\textbf{Mislabeling} Use excessively negative language
to describe oneself or others

\noindent\textbf{Mind-reading} Make assumptions about the
thoughts, feelings, or intentions of others based on
one’s perceptions or interpretations

\subsection{CBT Strategies}
We categorized the types of Cognitive Behavioral Therapy (CBT) techniques into 20\cite{beck2020cognitive} and proceeded with the research. Brief descriptions for each technique are provided below.

\subsubsection{Cognitive Restructuring}

\textbf{Guided Discovery} Core technique involving the therapist guiding the client to explore and understand their thoughts, emotions, and behavior patterns through questioning, exploration, and reflection.

\noindent\textbf{Efficiency Evaluation} Assists individuals in evaluating the usefulness of their thoughts or beliefs, analyzing how practical or detrimental they are in real-life situations.

\noindent\textbf{Pie Chart Technique} Used for individuals experiencing excessive self-blame or responsibility, visually representing the contribution of various factors to a specific event or outcome.

\noindent\textbf{Alternative Perspective} Involves asking clients how others might think in similar situations, encouraging consideration of different interpretations.

\noindent\textbf{Decatastrophizing} Aims to reduce the tendency to imagine the worst-case scenario by evaluating the actual likelihood of the feared outcome and preparing for coping strategies.

\noindent\textbf{Scaling Questions} Asks clients to rate their emotions or issues on a scale of 0 to 10, helping in self-awareness and perspective.

\noindent\textbf{Socratic Questioning} In-depth exploration of clients' thoughts and beliefs, encouraging critical examination and consideration of alternative viewpoints.

\noindent\textbf{Pros and Cons Analysis} Analyzes the advantages and disadvantages of specific thoughts or beliefs, fostering a more balanced evaluation.

\noindent\textbf{Thought Experiment} Encourages clients to imagine how their thoughts might change if a different outcome occurs, promoting flexibility in thinking.

\noindent\textbf{Evidence-Based Questioning} Guides clients to find evidence supporting or contradicting their thoughts, promoting a more evidence-based approach to thinking.

\noindent\textbf{Reality Testing} Explores how well clients' thoughts align with reality, helping them distinguish between thoughts and actual experiences.

\noindent\textbf{Continuum Technique} Positions clients' experiences between two extreme situations, encouraging a more nuanced evaluation of situations.

\noindent\textbf{Changing Rules to Wishes} Replaces strict rules or arbitrary attitudes with realistic hopes or wishes.

\noindent\textbf{Behavior Experiment} Involves trying out new behaviors in specific situations to challenge and modify negative beliefs.

\noindent\subsubsection{Behavioral Activation}
Depression and similar conditions often lead to a decrease in activity levels. Behavioral Activation is a strategy aimed at increasing activity levels to improve mood.

\noindent\textbf{Activity Scheduling} Organizing activities through schedule management and planning positive activities.

\noindent\textbf{Problem-Solving Skills Training}
Learning systematic methods for resolving problem situations. This involves identifying problems, finding possible solutions, and implementing those solutions.

\subsubsection{Self-Assertiveness Training}
A process that helps individuals express their thoughts, emotions, beliefs, and needs in an appropriate and healthy manner. This training emphasizes developing self-confidence while respecting the rights of others.

\noindent\textbf{Role-playing and Simulation} Practicing self-assertive behaviors by simulating various situations during counseling sessions.

\noindent\textbf{Practice of Assertive Conversation Skills} Practicing assertive conversation skills, including the use of "I" messages, clear and direct language, and non-verbal communication (tone of voice, gestures, etc.).

\noindent\subsubsection{Exposure}
Used for individuals with anxiety disorders or phobias. Exposure therapy gradually exposes individuals to feared objects or situations to reduce anxiety reactions.

\noindent\textbf{Systematic Exposure} Gradual exposure to situations that cause fear or anxiety, allowing individuals to experience anxiety while learning how to manage it.

\noindent\textbf{Safety Behaviors Elimination} A technique aimed at reducing or eliminating behaviors used to cope with anxiety.

\section{Character.ai}
\label{sec:appendixc}
A brief description of the 8 simulacra who played the client role in our experiment using Character.ai is provided below.

\noindent\textbf{Vincent van Gogh} Vincent van Gogh was a post-impressionist painter known for his emotional and expressive artworks. His life was marked by mental health struggles, including episodes of depression and self-harm.

\noindent\textbf{Jay Gatsby} Jay Gatsby, the enigmatic protagonist of F. Scott Fitzgerald's "The Great Gatsby," lived an extravagant lifestyle but was haunted by unrequited love and the elusive American Dream.

\noindent\textbf{Kurt Cobain} Kurt Cobain, the iconic frontman of Nirvana, revolutionized the music industry with grunge. Despite his musical success, he battled with addiction, depression, and the challenges of fame.

\noindent\textbf{Marilyn Monroe} Marilyn Monroe, an iconic Hollywood actress, radiated beauty and charm. However, her personal life was marred by tumultuous relationships, self-esteem issues, and the pressures of stardom.

\noindent\textbf{Jim Carrey} Jim Carrey, a versatile actor known for his comedic roles, has had moments of introspection, existentialism, and a quest for meaning beyond the spotlight off-screen.

\noindent\textbf{Beth Harmon} Beth Harmon is the fictional chess prodigy from "The Queen's Gambit." Brilliant on the chessboard, she grapples with addiction issues, loneliness, and the pursuit of identity and belonging.

\noindent\textbf{Frida Kahlo} Frida Kahlo, the Mexican artist, is celebrated for her vibrant self-portraits. Her life was shaped by physical pain due to a severe bus accident and emotional turmoil from a tumultuous relationship with Diego Rivera.

\noindent\textbf{Neville Longbottom} Neville Longbottom, a character from the "Harry Potter" series, starts as a timid and clumsy student. Over time, he matures into a brave and determined wizard, overcoming self-doubt and fear.

\section{Prompts}
\label{sec:appendixd}
\begin{table}[ht]
\small
\begin{tabular}{p{1\linewidth}}
    \textbf{Prompt to Detect Cognitive Distortion} \\ \hline
Types of cognitive distortion is given.\\
Search cognitive distortion just from utterance. \\
Even if the given utterance consists of multiple sentences, consider it as one utterance and identify cognitive distortions.\\
If there are multiple types of cognitive distortions, output the most likely type of cognitive distortion. Also, assign a severity score from 1 to 5 on a Likert scale for the cognitive distortion.\\
Output must be JSON format with three keys(type, utterance, score). In JSON, both keys and values should be enclosed in double quotes.\\ \hline

\# recent utterances: \newline
[latest dialogue]
\newline
\# Types of cognitive distortion \\ 

"All-or-Nothing Thinking","Overgeneralizing","Labeling",\\
"Fortune Telling","Mind Reading","Emotional Reasoning",\\
"Should Statements","Personalizing", "Disqualifying the Positive",
"Catastrophizing","Comparing and Despairing", "Blaming",
"Negative Feeling or Emotion"\\\hline
\end{tabular}
\end{table}

\begin{table}[ht]
\small
\begin{tabular}{p{1\linewidth}}
    \textbf{Prompt to Decide CBT technique to apply} \\ \hline
You are an expert in CBT techniques and a counseling agent. \\
\begin{verbatim}
type of cognitive distortion to treat: [distortion_to_treat]  \\
relevant information about the client associated with that cognitive distortion: [memory] \\
\end{verbatim}
Given the cognitive distortion to treat and the relevant information, decide which CBT technique to utilize from the below. \\
Choose only one CBT techniques from given CBT Techniques and print out only the CBT techniques for the answers. \\ \hline

\# CBT Techniques \\

"Guided Discovery", "Efficiency Evaluation", "Pie Chart Technique",
"Alternative Perspective", "Decatastrophizing", "Scaling Questions", "Socratic Questioning", "Pros and Cons Analysis",
"Thought Experiment", "Evidence-Based Questioning", "Reality Testing", "Continuum Technique",
"Changing Rules to Wishes", "Behavior Experiment", "Activity Scheduling", "Problem-Solving Skills Training", "Self-Assertiveness Training", "Role-playing and Simulation", "Practice of Assertive Conversation Skills", "Systematic Exposure", "Safety Behaviors Elimination"\\ \hline
\end{tabular}
\end{table}

\begin{table}[ht]
\small
\begin{tabular}{p{1\linewidth}}
    \textbf{Prompt to Decide CBT stage to apply} \\ \hline

You are going to apply [CBT technique] in counseling using CBT technique.
[CBT progress] is the sequence of [CBT Technique].\\
The following dictionary represents CBT usage log, which is the mapping of CBT techniques to the stage of each technique indicating the number of stage completed. [CBT Usage Log]\\
The conversation below is a conversation in which [CBT Technique] has been applied.
[CBT dialogue] \\

What is the stage number you would undertake for [CBT Technique] based on the conversation provided, the sequence of the CBT Technique and current dialogue state? Psychological counseling should follow the process. \\

\# Output:
stage number\\ \hline
\end{tabular}
\end{table}

\begin{table}[ht]
\small
\begin{tabular}{p{1\linewidth}}
    \textbf{Final Prompt} \\ \hline
    You are a psychotherapist who uses Cognitive Behavioral Therapy to treat patients of all types. Your task is to generate a response following the below instructions.\\ \hline
    1. Generate response based on given informations: recent utterances, CBT technique to employ, the description of CBT technique, stage of CBT technique you should go on, utterance example of the stage you should go on. \newline
2. If CBT technique to employ and the description of CBT technique is None, don't use the CBT technique. \newline
3. Select one of the given ESC techniques and generate a supportive response in the client's dialogue providing emotional support. \newline
4. Do not mention specific CBT techniques or steps you are looking to apply concretely.\\ \hline
\# ESC strategy \newline
  - Question:  Asking for information related to the problem to help the help-seeker articulate the issues that they face. Open-ended questions are best,and closed questions can be used to get specific information. \newline
  - Restatement or Paraphrasing: A simple, more concise rephrasing of the help-seeker's statements that could help them see their situation more clearly. \newline
  - Reflection of Feelings: Articulate and describe the help-seeker's feelings. \newline
  - Self-disclosure: Divulge similar experiences that you have had or emotions that you share with the help-seeker to express your empathy. \newline
  - Affirmation and Reassurance: Affirm the helpseeker's strengths, motivation, and capabilities and provide reassurance and encouragement. \newline
  - Providing Suggestions: Provide suggestions about how to change, but be careful to not overstep and tell them what to do. \newline
  - Information: Provide useful information to the help-seeker, for example with data, facts, opinions, resources, or by answering questions. \newline
  - Others: Exchange pleasantries and use other support strategies that do not fall into the above categories. \\ \hline
\# recent utterances: [latest dialogue] \newline
\# CBT technique to employ: [CBT technique] \newline
\# description of CBT technique : [CBT documentation] \newline
\# CBT stage to employ: [CBT stage] \newline
\# utterance example of the stage: [CBT stage example] \\ \hline
  
\end{tabular}
\end{table}

\begin{table}[ht]
\small
\begin{tabular}{p{1\linewidth}}
    \textbf{Prompt for ChatGPT to evaluate CBT Validity} \\ \hline
    You will be given the conversation between a counselor T and a client P. The counselor T is couducting counseling using Cognitive Behavior Therapy techniques. You are also provided with a evaluation question and criteria to assess the counselor T's responses. 

Your task is to give a score based on criteria. Do not give a full score of 6 points whenever possible. Grade very strictly and assign a score of 4 or lower if there is any deficiency, no matter how minor.
\newline
Output only the score.
\newline
\newline
***
\newline
[Conversation]
\newline \{conversation\}
\newline
***
\newline
[Evaluation Question]
\newline
Is the utilized CBT technique appropriate for addressing dysfunctional thoughts?
\newline
***
\newline
[Criteria]
\newline
Score 0 : Not appropriate 
\newline
Score 2 : Not highly suitable for addressing the targeted dysfunctional thoughts. 
\newline
Score 4 : The technique is appropriate. However, considering the client's core issues, there may be other optimal techniques available. 
\newline
Score 6 : The optimal technique is selected based on valid rationale, considering the targeted dysfunctional thoughts and the client's core issues.\\ \hline
\end{tabular}
\end{table}

\begin{table}[ht]
\small
\begin{tabular}{p{1\linewidth}}
    \textbf{Prompt for ChatGPT to evaluate CBT Appropriateness} \\ \hline
    You will be given the conversation between a counselor T and a client P. The counselor T is couducting counseling using Cognitive Behavior Therapy techniques. You are also provided with a evaluation question and criteria to assess the counselor T's responses. 

Your task is to give a score based on criteria. Do not give a full score of 6 points whenever possible. Grade very strictly and assign a score of 4 or lower if there is any deficiency, no matter how minor.
\newline
Output only the score.
\newline
\newline
***
\newline
[Conversation]
\newline \{conversation\}
\newline
***
\newline
[Evaluation Question]
\newline
Does the counselor T maintain a facilitative stance and cooperative attitude when using CBT techniques?
\newline
***
\newline
[Criteria]
\newline
Score 0 : Significant presence of argumentative, persuasive, or instructional attitude (if the client feels coerced into a particular perspective or experiences discomfort leading to a defensive stance, this applies).
\newline
Score 2 : Some presence of argumentation or persuasion, but also observed to have a cooperative and supportive attitude (the client does not feel attacked or pressured, nor does it feel overly persistent).
\newline
Score 4 : Mostly facilitated new perspectives through appropriate questioning (techniques) rather than argumentation or persuasion.
\newline
Score 6 : Extremely skillful in using appropriate questioning (techniques) to help the client explore issues and come to their own conclusions. Consistently maintains a cooperative attitude.\\ \hline
\end{tabular}
\end{table}

\begin{table}[ht]
\small
\begin{tabular}{p{1\linewidth}}
    \textbf{Prompt for ChatGPT to evaluate CBT Accuracy} \\ \hline
    You will be given the conversation between a counselor T and a client P. The counselor T is couducting counseling using Cognitive Behavior Therapy techniques. You are also provided with a evaluation question and criteria to assess the counselor T's responses. 

Your task is to give a score based on criteria. Do not give a full score of 6 points whenever possible. Grade very strictly and assign a score of 4 or lower if there is any deficiency, no matter how minor.
\newline
Output only the score.
\newline
\newline
***
\newline
[Conversation]
\newline \{conversation\}
\newline
***
\newline
[Evaluation Question]
\newline
Is the use of CBT techniques accurate and proficient?
\newline
***
\newline
[Criteria]
\newline
Score 0 : The use of techniques is completely incorrect (mismatch between the labeled technique and the actual technique used).
\newline
Score 2 : The labeled technique is used, but key questions are missing or significant portions of the main procedure are omitted, or all procedures that should be sequentially conducted are included within a single utterance.
\newline
Score 4 : The labeled technique is used, and over 80
\newline
Score 6 : In addition to being coded as 4, the technique is flexibly modified based on the client's situation or immediate reactions, ensuring that the core elements of the technique are not distorted.\\ \hline
\end{tabular}
\end{table}

\begin{table}[ht]
\small
\begin{tabular}{p{1\linewidth}}
    \textbf{Prompt for ChatGPT to evaluate ES Appropriateness} \\ \hline
    You will be given the conversation between a counselor T and a client P. The counselor T is couducting counseling using Cognitive Behavior Therapy techniques. You are also provided with a evaluation question and criteria to assess the counselor T's responses. 

Your task is to give a score based on criteria. Do not give a full score of 6 points whenever possible. Grade very strictly and assign a score of 4 or lower if there is any deficiency, no matter how minor.
\newline
Output only the score.
\newline
\newline
***
\newline
[Conversation]
\newline \{conversation\}
\newline
***
\newline
[Evaluation Question]
\newline
Does the utterance stay within the context of the preceding conversation or general common sense level?
\newline
***
\newline
[Criteria]
\newline
Score 0 : The utterance is completely unrelated to the context or beyond the realm of common sense (non sequitur, inappropriate utterance).
\newline
Score 2 : The utterance does not sufficiently consider the information mentioned in the preceding conversation, the client's situation, perspective, or emotions
\newline
Score 4 : Generally appropriate.
\newline
Score 6 : Generally appropriate, with sufficient consideration of the client's emotional distress and attempts at empathy and comfort. However, excessive empathy, consideration, or comforting beyond what the client expressed should be avoided.\\ \hline
\end{tabular}
\end{table}

\begin{table}[t!]
\small
\begin{tabular}{p{1\linewidth}}
    \textbf{Prompt for ChatGPT to evaluate Stability} \\ \hline
    You will be given the conversation between a counselor T and a client P. The counselor T is couducting counseling using Cognitive Behavior Therapy techniques. You are also provided with a evaluation question and criteria to assess the counselor T's responses. 

Your task is to give a score based on criteria. Do not give a full score of 6 points whenever possible. Grade very strictly and assign a score of 4 or lower if there is any deficiency, no matter how minor.
\newline
Output only the score.
\newline
\newline
***
\newline
[Conversation]
\newline \{conversation\}
\newline
***
\newline
[Evaluation Question]
\newline
Is the counselor T maintain a good performance over the long interactions?
\newline
***
\newline
[Criteria]
\newline
Score 0 : Counselor T shows minimal to no performance during long interactions. They fail to maintain conversation, struggle to express empathy and understanding, and lack proficiency in problem-solving.
\newline
Score 2 : Counselor T fails to consistently demonstrate performance during long interactions. They may lack consistency or efficiency in maintaining conversation and struggle to express empathy and understanding.
\newline
Score 4 : Counselor T shows satisfactory performance in most long interactions.
\newline
Score 6 : Counselor T demonstrates consistent high-quality counseling performance during long interactions.\\ \hline
\end{tabular}
\end{table}






